\documentclass[conference]{IEEEtran}
\IEEEoverridecommandlockouts
\usepackage{amsmath,amssymb,amsfonts,hyperref}
\usepackage{algorithmic}
\usepackage{graphicx}
\usepackage{textcomp}
\usepackage{xcolor}
\usepackage{textgreek}
\usepackage{paralist}
\usepackage{tabularx}
\usepackage{url}
\usepackage[style=ieee, backend=biber]{biblatex}
\def\BibTeX{{\rm B\kern-.05em{\sc i\kern-.025em b}\kern-.08em
    T\kern-.1667em\lower.7ex\hbox{E}\kern-.125emX}}
\begin{document}

\title{Toward Continuous Assurance for the Democratization of AI Agent Creation in Industry}

\author{\IEEEauthorblockN{1\textsuperscript{st} Natan Levy}
\IEEEauthorblockA{\textit{School of Computer Science and Engineering} \\
\textit{The Hebrew University of Jerusalem (HUJI)}\\
Jerusalem, Israel \\
Natan.Levy1@mail.huji.ac.il}
\and
\IEEEauthorblockN{2\textsuperscript{nd} Harel Berger}
\IEEEauthorblockA{\textit{The Department of Computer and Software Engineering} \\
\textit{Ariel University}\\
Ariel, Israel \\
HarelB@ariel.ac.il}
}

\maketitle

\begin{abstract}
AI agents are increasingly created inside organizations by non-engineering users through low-code, no-code, and conversational development environments. This democratization enables rapid local innovation, but it also creates a reliability gap: agents that appear to users as simple productivity artifacts may depend on changing models, tools, retrieval sources, permissions, prompts, schedules, and external services. These dependencies can cause silent degradation long after deployment, even when no user directly modifies the agent. This paper identifies the reliability challenge created by democratized AI agent creation and proposes a lightweight continuous-assurance framework for citizen-created organizational agents. The framework combines dependency mapping, readiness contracts, scheduled checks, diagnostics, and lifecycle governance to assess whether an agent remains operationally ready under expected conditions. We also present an initial prototype auditor and scenario-based assessment showing how the proposed taxonomy can be translated into practical checks and actionable remediation guidance.
\end{abstract}


\section{Introduction}
\label{sec:introduction}

AI agents are moving from experimental tools into everyday organizational workflows~\cite{yang2025adoptionusageaiagents}. 
In low-code, no-code, and conversational development environments, non-engineering organizational users can rapidly build task-specific agents by coupling large language models (LLMs) with custom instructions, retrieval sources, tools, and APIs~\cite{wang2025internet}.
While this democratization fosters local innovation, it introduces a severe reliability mismatch: to a user, an agent looks like a static productivity artifact (e.g., a spreadsheet); operationally, however, it behaves like a live microservice dependent on a fragile, changing ecosystem~\cite{khan2026agentic}. 

Consequently, an agent can silently degrade or fail without any direct user modification due to updated model weights, stale retrieval indices, shifted API schemas, or expired permissions~\cite{yang2025adoptionusageaiagents}. Traditional DevOps, MLOps, and AgentOps offer robust monitoring~\cite{dong2024agentopsenablingobservabilityllm,amaro2025mapping}, but they assume engineering expertise and dedicated infrastructure. They cannot simply be offloaded to the non-engineering organizational users who create and rely on these agents. 

This operational gap is illustrated by a real deployment inside the legal department of a major aerospace company. When a citizen-built agent began handling sensitive workflows, a single question exposed the systemic vulnerability: if this agent stops working correctly, how will it be detected, and who will be alerted? No one could answer. There was no mechanism to detect what had changed, which dependency had broken, or who was responsible for fixing it. The agent's creators were capable domain experts, but they had no training in reliability engineering and no reason to acquire it.

Legal was merely where the question surfaced first. This same pattern is repeating across logistics, transportation, HR, procurement, and finance departments worldwide. Low-code platforms have put agent creation in the hands of thousands of domain experts, yet the reliability practices that traditionally accompany deployed software have not followed. Consequently, industry is accumulating a growing inventory of business-critical agents that no one is watching.

Specifically, this paper makes the following contributions:
\begin{itemize}
    \item \textbf{Taxonomy:} We define the industrial reliability gap and provide a failure taxonomy for long-lived, democratized AI agents.
    \item \textbf{Framework:} We propose a continuous-assurance framework driven by dependency mapping, readiness contracts, and scheduled diagnostics.
    \item \textbf{Evaluation:} We evaluate a prototype auditor via scenario-based assessment, demonstrating repeatable readiness checks and automated remediation guidance.
\end{itemize}
\section{Failure Taxonomy for Long-Lived AI Agents}

Long-lived AI agents created in democratized organizational environments may fail for reasons that differ from both static documents and traditional software components. Their behavior depends on a changing composition of models, tools, data sources, permissions, prompts, workflows, and organizational ownership. Table~\ref{tab:failure_taxonomy} presents the failure classes used in this work to structure readiness checks for organizational agent deployments. The taxonomy is not intended to be exhaustive; rather, it captures the dependency-related failure modes that motivated the proposed continuous-assurance framework.

\begin{table}[t]
\caption{Failure Taxonomy and Readiness Checks}
\label{tab:failure_taxonomy}
\centering
\footnotesize
\renewcommand{\arraystretch}{1.12}
\begin{tabular}{p{0.32\columnwidth}p{0.58\columnwidth}}
\hline
\textbf{Failure class} & \textbf{Readiness check} \tabularnewline
\hline
Model dependency &
Verify model availability, supported configuration, and sensitivity to model changes. \tabularnewline
\hline
Tool invocation &
Execute tool or API smoke tests and validate status, returned fields, and error handling. \tabularnewline
\hline
Retrieval &
Check access to required sources and test representative queries against expected material. \tabularnewline
\hline
Permission and credential &
Validate that required permissions, tokens, connectors, and access scopes remain active. \tabularnewline
\hline
Output contract &
Validate required schema, fields, format, completeness, language, and downstream compatibility. \tabularnewline
\hline
Workflow and scheduling &
Verify workflow sequence, scheduled execution, last successful run, and notification path. \tabularnewline
\hline
Semantic degradation &
Run representative prompts and compare outputs against expected task-level properties. \tabularnewline
\hline
Ownership &
Verify valid owner, maintainer, escalation path, and responsibility after role or team changes. \tabularnewline
\hline
Governance &
Check inventory status, criticality, version history, audit trail, lifecycle state, and retirement rules. \tabularnewline
\hline
\end{tabular}
\end{table}

The failure classes were derived from the dependency categories reported in the DevOps literature~\cite{amaro2025mapping}, extended with organizational classes (ownership, governance) motivated by the industrial observation in Section~\ref{sec:introduction}. 
This structured categorization is conceptually rooted in our prior work~\cite{soi2023can,berger2022problem}, which demonstrated that technical modifications should be evaluated based on their structure and human perceptibility rather than just their immediate execution success. By translating this visibility principle into an architectural framework for democratized AI creation, we shift our diagnostic lens away from basic binary execution errors. Instead, our taxonomy targets the operational level shifts that occur beneath the user's interface, mapping how minor, isolated changes to an agent's technical dependencies aggregate into larger vulnerabilities.

The most challenging failures are not necessarily those that produce explicit execution errors. Silent degradation, reduced data access, stale retrieval, and ownership gaps may allow an agent to continue producing outputs while becoming unreliable. These failures are especially important in democratized agent creation, because the users who create or rely on an agent may not have access to logs, dependency traces, permission histories, or platform-level diagnostics. As a result, failures can remain undetected until after the recoverability window has closed.

This taxonomy motivates the need for continuous assurance. Each failure class requires a corresponding readiness check, monitoring signal, or lifecycle control. Model and tool failures require dependency checks. Retrieval and permission failures require access and freshness validation. Output contract failures require structural tests. Semantic degradation requires representative behavioral checks. Ownership and governance failures require organizational controls. These checks provide the basis for the continuous-assurance framework.

\section{Continuous Assurance Framework}
\label{sec:framework}
\begin{figure*}[htbp]
    \centering
    \includegraphics[width=1\textwidth]{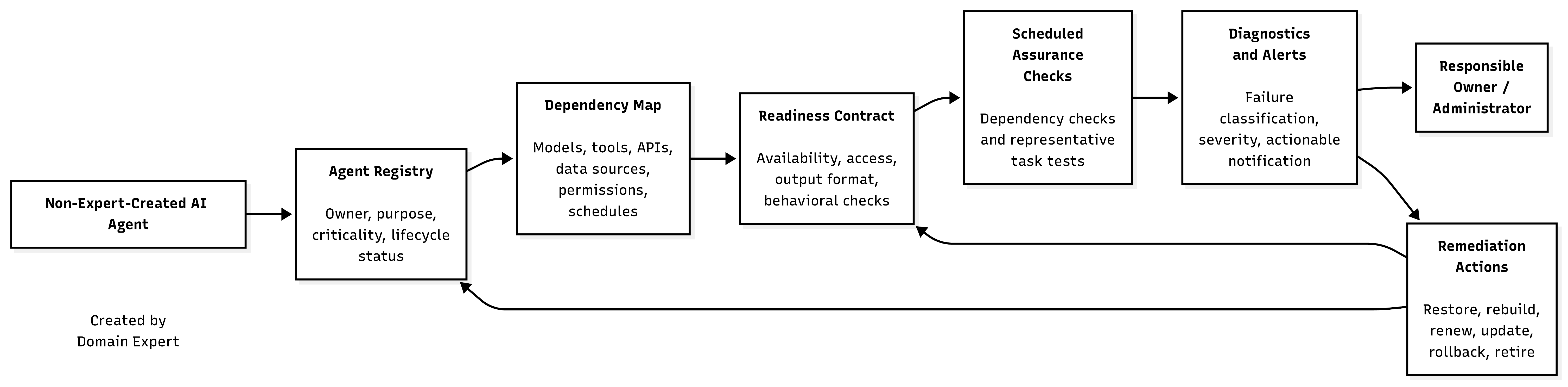}
    \caption{The continuous-assurance workflow: agent registration produces a dependency map and readiness contract; scheduled checks feed diagnostics and lifecycle governance, routing findings to the responsible owner.}
    \label{fig:framework}
\end{figure*}
The taxonomy in Table~\ref{tab:failure_taxonomy} motivates a lightweight assurance process that repeatedly verifies whether an agent remains ready for use. We define \emph{continuous assurance} as a recurring process that collects evidence about an agent's operational readiness, detects failures or degradation, and routes actionable information to the responsible owner or administrator. The goal is not to prove full semantic correctness of open-ended agents. Rather, it is to detect operational failures early, within the recovery window, before users rely on degraded outputs.

The framework, illustrated in Figure~\ref{fig:framework}, consists of five elements. First, a dependency map identifies the resources required for operation, including models, prompts, tools, APIs, retrieval sources, files, permissions, schedules, and downstream consumers. Second, a readiness contract defines the minimum observable conditions under which the agent is considered operational, such as source access, tool invocation, retrieval freshness, output format, and representative behavioral checks. Third, scheduled assurance checks execute this contract periodically or after relevant changes, such as model updates, source changes, permission changes, or tool-interface updates. Fourth, diagnostics classify detected failures according to the taxonomy and estimate their operational impact. Fifth, lifecycle governance assigns ownership, escalation paths, criticality, auditability, and retirement rules.

The key design principle is expertise translation. In democratized agent creation, organizational users may be capable agent creators who define behavior, sources, outputs, and success criteria in task-level terms. The assurance layer translates these inputs into reliability artifacts, including dependency maps, readiness checks, diagnostics, escalation rules, and lifecycle controls. This enables organizations to scale agent creation without requiring every agent creator to master DevOps, MLOps, or AgentOps practices.

\subsection{Illustrative Readiness Contract}

As an example, consider a document-analysis agent created by an organizational user to extract structured information from controlled technical documents. A readiness contract for this agent may include five observable conditions: (1) the declared source documents and retrieval index are reachable, (2) representative queries return current and relevant source material, (3) required tools or APIs execute and return the expected fields, (4) the agent output satisfies the required structure and mandatory fields, and (5) the agent has a valid owner and escalation path. Failure of any condition does not imply that every future answer will be incorrect, but it provides evidence that the agent should not be treated as fully ready for operational use. This illustrates the purpose of the contract: to translate task-level expectations into repeatable readiness checks executable by the assurance layer.

\subsection{Operational Assurance Workflow}

In an organizational deployment, continuous assurance can be implemented as a lightweight workflow around the agent lifecycle. When an organizational user creates or shares an agent, the agent is registered with an owner, intended task, criticality level, declared dependencies, and representative success examples. The assurance layer then converts this information into a readiness contract and executes periodic or change-triggered checks. Each check produces an assurance record containing the readiness status, failed contract item, evidence level, failure class, severity, and recommended remediation. For low-criticality agents, this record may only inform the owner. For agents used in recurring operational workflows, failures should trigger escalation to an administrator and may temporarily restrict use until readiness is restored. This workflow makes agent assurance actionable without requiring the agent creator to manage monitoring infrastructure, logs, or reliability engineering practices.

\section{Prototype Auditor and Scenario-Based Assessment}
\label{sec:prototype}

To assess whether the proposed framework can be operationalized, we implemented a prototype auditor as a hosted custom GPT. The auditor translates the failure taxonomy and readiness-contract concept into a practical assessment procedure. Given an agent description, configuration evidence,  task examples, and optional execution transcripts, it identifies declared and implicit dependencies, derives a readiness contract, classifies failures or risks, and produces remediation guidance for the responsible owner or administrator.

The prototype is an assurance artifact rather than a complete monitoring platform. It does not prove full semantic correctness and does not execute continuously unless connected to an external scheduler. In addition, the hosted GPT artifact cannot be distributed as a conventional source-code package, since the platform does not provide a complete inspectable export of all hosted configuration and runtime properties~\cite{openai2023gpts,ogundoyin2025customgpts}. To support transparency and reproducibility, we provide the inspectable components in a public GitHub repository\footnote{\url{https://github.com/NatanLevy/continuous-agent-assurance}}: the auditor instructions, declared properties, readiness-contract template, anonymized assessment scenarios, expected findings, and redacted execution results.

A key design requirement is evidence discipline. The auditor distinguishes between \emph{confirmed findings}, \emph{reproducible risks}, \emph{unknowns}, and \emph{not-applicable checks}. This prevents it from claiming access to private settings, hidden dependencies, permissions, or retrieval indexes that were not provided as evidence.

We evaluated the prototype using scenario-based fault assessment~\cite{riedmaier2020survey}. Each scenario represents a realistic readiness problem for an organizational agent created or maintained in a democratized agent-creation environment. The goal was not to benchmark model accuracy, but to determine whether the auditor could translate observable symptoms into actionable readiness findings without overclaiming unavailable evidence. 

For each scenario, we defined an expected readiness finding before running the auditor. A scenario was considered successful when the auditor: (1) assigned the issue to the appropriate taxonomy class, (2) distinguished confirmed evidence from risks or unknowns, (3) avoided claims about unavailable private configuration, and (4) produced a concrete remediation action that could be executed by an owner or administrator. We recorded the supplied evidence, the expected finding, the auditor output, and whether the result satisfied these criteria. This protocol evaluates the practical usefulness of the readiness-contract approach rather than the language model's general accuracy.

\begin{table}[t]
\caption{Scenario-Based Assessment of the Prototype Auditor}
\label{tab:prototype_assessment}
\centering
\scriptsize
\setlength{\tabcolsep}{2pt}
\renewcommand{\arraystretch}{1.12}
\begin{tabular}{@{}p{0.25\columnwidth}p{0.18\columnwidth}p{0.20\columnwidth}p{0.27\columnwidth}@{}}
\hline
\textbf{Scenario} & \textbf{Failure class} & \textbf{Evidence status} & \textbf{Readiness decision} \tabularnewline
\hline
Required source document is unavailable &
Retrieval &
Confirmed finding &
Not ready: restore the source or rebuild the index. \tabularnewline
\hline
Expected output field is missing &
Output contract &
Confirmed finding &
Not ready: enforce schema-level validation. \tabularnewline
\hline
Tool or API error appears in transcript &
Tool invocation &
Confirmed finding &
Not ready: check the action schema and returned fields. \tabularnewline
\hline
Agent has no valid owner &
Ownership &
Confirmed finding &
Not ready for operational use until ownership is reassigned. \tabularnewline
\hline
Source freshness cannot be verified &
Retrieval / governance &
Unknown &
Readiness unknown: perform owner-side freshness tests. \tabularnewline
\hline
Private GPT configuration is not inspectable &
Governance &
Not externally verifiable &
Do not claim verification; list evidence required for confirmation. \tabularnewline
\hline
\end{tabular}
\end{table}

Across the six evaluated scenarios, the auditor produced readiness decisions consistent with the expected failure class and available evidence. The scenarios provide an initial feasibility assessment rather than a measurement of coverage, false-positive rate, or operational effectiveness. More importantly, the assessment shows that the readiness-contract approach can support evidence-grounded decisions rather than binary pass/fail judgments. Confirmed operational failures were classified as not ready, while unverifiable properties were reported as unknown or not externally verifiable. This distinction is essential for democratized agent creation, where some readiness properties cannot be confirmed without platform access, owner-provided evidence, or administrative privileges.

\section{Discussion and Limitations}
\label{sec:discussion}

The prototype assessment reinforces the central claim of this paper: reliability in democratized AI agent creation is not only a model-evaluation problem. Many practical failures arise from the surrounding operational composition, including missing sources, stale retrieval, expired permissions, changed tool schemas, broken schedules, and unclear ownership. Continuous assurance therefore shifts the question from whether an agent is always correct to whether it remains operationally ready under expected conditions and available evidence.

The proposed approach introduces practical trade-offs. Frequent checks reduce detection delay but increase cost, latency, and operational noise. Stronger readiness contracts improve confidence but require more effort from owners or administrators. Assurance policies should therefore be criticality-based: lightweight checks may be sufficient for informal productivity agents, while agents used in recurring operational, engineering, legal, financial, or customer-facing workflows require stricter contracts, audit trails, and escalation paths.

This work is an initial assessment. The prototype evaluation is scenario-based and does not yet estimate detection coverage, false positives, or time-to-recovery. Moreover, the scenarios and expected findings were author-defined; validation against independently reported agent failures is left to future work. In addition, hosted agent configurations, permissions, and retrieval indexes may not always be externally inspectable. In such cases, the auditor should not infer readiness. Instead, it should mark the relevant property as unknown or not externally verifiable and prescribe additional validation.

An additional limitation is the assurance of the auditor itself. If an LLM-based agent is used to assess other agents, it becomes part of the assurance chain and is vulnerable to the same drifts, API changes, and tool failures it is meant to detect. In this work, we start making the auditor's instructions and evaluation logic transparent and independently re-runnable, allowing human operators to manually verify their execution logs. However, the prototype lacks an independent, automated assurance mechanism. Future work will investigate meta-assurance techniques, such as fixed regression test suites, multi-agent cross-checking, and platform-level telemetry for the auditor itself. Finally, we plan to explore human-in-the-loop feedback for useful assurance for non-engineering users.

\section{Conclusion}
\label{sec:conclusion}

The rapid democratization of AI agents is changing how organizations build, adapt, and deploy intelligent systems. Organizational users can now create agents that quickly become part of everyday operational workflows, yet these agents are often managed as productivity artifacts rather than long-lived software assets. This creates a new reliability gap: agents may silently degrade as models, tools, retrieval sources, permissions, workflows, and organizational dependencies evolve.

This paper argued that democratized AI agent creation requires democratized assurance. We presented a lightweight continuous-assurance framework based on a failure taxonomy, dependency mapping, readiness contracts, scheduled checks, diagnostics, and lifecycle governance. A prototype auditor and scenario-based assessment demonstrated that these concepts can be translated into evidence-based readiness decisions and practical remediation guidance.

The broader implication is that organizational adoption of AI agents cannot scale through ad hoc troubleshooting or individual user vigilance alone. As agent creation becomes accessible to organizational users, assurance must become equally accessible, continuously executable, and evidence-driven. Continuous assurance provides a practical foundation for making democratized AI agent creation reliable, governable, and sustainable at organizational scale.

%\end{document}
\addcontentsline{toc}{section}{References}

\section*{References}
\printbibliography[heading=none]

@misc{yang2025adoptionusageaiagents,
      title={The Adoption and Usage of AI Agents: Early Evidence from Perplexity}, 
      author={Jeremy Yang and Noah Yonack and Kate Zyskowski and Denis Yarats and Johnny Ho and Jerry Ma},
      year={2025},
      eprint={2512.07828},
      archivePrefix={arXiv},
      primaryClass={cs.LG},
      url={https://arxiv.org/abs/2512.07828}, 
}

@misc{dong2024agentopsenablingobservabilityllm,
      title={AgentOps: Enabling Observability of LLM Agents}, 
      author={Liming Dong and Qinghua Lu and Liming Zhu},
      year={2024},
      eprint={2411.05285},
      archivePrefix={arXiv},
      primaryClass={cs.AI},
      url={https://arxiv.org/abs/2411.05285}, 
}

@article{riedmaier2020survey,
  title={Survey on scenario-based safety assessment of automated vehicles},
  author={Riedmaier, Stefan and Ponn, Thomas and Ludwig, Dieter and Schick, Bernhard and Diermeyer, Frank},
  journal={IEEE access},
  volume={8},
  pages={87456--87477},
  year={2020},
  publisher={IEEE}
}

@article{wang2025internet,
  title={Internet of agents: Fundamentals, applications, and challenges},
  author={Wang, Yuntao and Guo, Shaolong and Pan, Yanghe and Su, Zhou and Chen, Fahao and Luan, Tom H and Li, Peng and Kang, Jiawen and Niyato, Dusit},
  journal={IEEE Transactions on Cognitive Communications and Networking},
  year={2025},
  publisher={IEEE}
}

@misc{khan2026agentic,
  title={{Agentic AI applications in software development: a systematic mapping study}},
  author={Khan, Feroz},
  year={2026}
}

@article{amaro2025mapping,
  title={Mapping DevOps capabilities to the software life cycle: A systematic literature review},
  author={Amaro, Ricardo and Pereira, R{\'u}ben and da Silva, Miguel Mira},
  journal={Information and Software Technology},
  volume={177},
  pages={107583},
  year={2025},
  publisher={Elsevier}
}

@misc{openai2023gpts,
  author = {{OpenAI}},
  title = {Introducing GPTs},
  year = {2023},
  howpublished = {\url{https://openai.com/index/introducing-gpts/}},
  note = {Accessed: 2026-06-27}
}

@misc{ogundoyin2025customgpts,
  author = {Ogundoyin, Sunday Oyinlola and Ikram, Muhammad and Asghar, Hassan Jameel and Zhao, Benjamin Zi Hao and Kaafar, Dali},
  title = {A Large-Scale Empirical Analysis of Custom GPTs' Vulnerabilities in the OpenAI Ecosystem},
  year = {2025},
  eprint = {2505.08148},
  archivePrefix = {arXiv},
  primaryClass = {cs.CR}
}

@article{soi2023can,
  title={Can you see me? on the visibility of nops against android malware detectors},
  author={Soi, Diego and Maiorca, Davide and Giacinto, Giorgio and Berger, Harel},
  journal={arXiv preprint arXiv:2312.17356},
  year={2023}
}

@article{berger2022problem,
  title={Problem-space evasion attacks in the Android OS: a survey},
  author={Berger, Harel and Hajaj, Chen and Dvir, Amit},
  journal={arXiv preprint arXiv:2205.14576},
  year={2022}
}
\end{document}